\documentclass[10pt,twocolumn,letterpaper]{article}

\usepackage[pagenumbers]{cvpr}      

\usepackage{graphicx}
\usepackage{amsmath}
\usepackage{amssymb}
\usepackage{comment}
\usepackage{booktabs}
\usepackage{multicol}
\usepackage{adjustbox}
\usepackage{dblfloatfix}
\usepackage[pagebackref,breaklinks,colorlinks]{hyperref}

\usepackage[capitalize]{cleveref}
\crefname{section}{Sec.}{Secs.}
\Crefname{section}{Section}{Sections}
\Crefname{table}{Table}{Tables}
\crefname{table}{Tab.}{Tabs.}

\def\cvprPaperID{4699} 
\def\confName{CVPR}
\def\confYear{2022}

\begin{document}

\title{Fair Tree Classifier using Strong Demographic Parity} 

\author{
António Pereira Barata$^{1,*}$ \and
Frank W. Takes$^1$ \and
H. Jaap van den Herik$^2$ \and
Cor J. Veenman$^1$ \and


\small $^1$Leiden Institute of Advanced Computer Science, Leiden University, the Netherlands\\
\small $^2$Leiden Centre of Data Science, Leiden University, the Netherlands \and
\small $^*$Corresponding author\\\small\texttt{apbarata@gmail.com}
}
\maketitle

\begin{abstract}
When dealing with sensitive data in automated data-driven decision-making, an important concern is to learn predictors with high performance towards a class label, whilst minimising for the discrimination towards any sensitive attribute, like gender or race, induced from biased data.
A few hybrid tree optimisation criteria exist that combine classification performance and fairness.
Although the threshold-free ROC-AUC is the standard for measuring traditional classification model performance, current fair tree classification methods mainly optimise for a fixed threshold on both the classification task as well as the fairness metric.
In this paper, we propose a compound splitting criterion which combines threshold-free (i.e., strong) demographic parity with ROC-AUC termed SCAFF ---Splitting Criterion AUC for Fairness--- and easily extends to bagged and boosted tree frameworks.
Our method simultaneously leverages multiple sensitive attributes of which the values may be multicategorical or intersectional, and is tunable with respect to the unavoidable performance-fairness trade-off.
In our experiments, we demonstrate how SCAFF generates models with performance and fairness with respect to binary, multicategorical, and multiple sensitive attributes.
\end{abstract}

\section{Introduction}\label{intro}
The application of machine learning algorithms for classification has become ubiquitous within an abundance of domains~\cite{brink2016real,sarker2021machine}.
Great dependency on automated decision-making, however, gives rise to concerns over model discrimination;
e.g., bias was reported by Amazon's automatic recruitment tool in which women unfairly scored lower.
It turns out that models were trained on resumes submitted mostly by men, thus disadvantaging women a priori~\cite{reuters2018}.
To prevent the modelling of historical biases, it is of the utmost importance to develop fairness-aware methods~\cite{proposal21ec}.

A fair classification model has mainly two goals:
(1) to make adequate class predictions from \textit{unseen} observations; and
(2) to ensure that those class predictions are as independent of a sensitive attribute as possible~\cite{NEURIPS2020_d0921d44, NEURIPS2020_ac3870fc}.
\linebreak
In addition, the \textit{performance-fairness trade-off} ---the phenomenon in which the lesser the fairness of an algorithm, the greater its predictive capabilities and vice-versa~\cite{kleinberg2016inherent}--- should be tunable to satisfy the ethical, legal, and societal needs of the end user.
Such a classifier is most commonly learned by jointly optimising towards a classification performance measure and a fairness measure.
Traditionally, fairness measures such as demographic parity~\cite{dwork2012fairness}, equal opportunity~\cite{corbett2018measure}, or equalised odds~\cite{hardt2016equality} are used.
These fairness measures are all threshold-dependent.

Considering a classification model with continuous output, a decision threshold must be set to produce class predictions, upon which those measures are reliant.
In other words, fairness would only be ensured with respect to that particular threshold.
To counter this limitation, the threshold-independent fairness measure termed \textit{strong demographic parity} was proposed in~\cite{jiang2020wasserstein}.
It extends the aforementioned demographic parity by considering fairness throughout the entire range of possible decision thresholds.
The authors only considered a logistic regression classifier implementation.

Tree-based algorithms are still regarded as a state-of-the-art solution~\cite{zabihi2017detection, dogru2018traffic, angenent2020large}.
The prevalence of tree-based approaches in the literature is mostly due to
(1) model interpretability,
(2) their tendency to not overfit when used as ensembles,
(3) requiring little data pre-processing, and
(4) handling mixed data types and missingness~\cite{dogru2018traffic}.
Past work on tree splitting criteria has shown positive results with respect to threshold-dependent fairness~\cite{kamiran2010discrimination}.
Hence, there is need to extend it towards the threshold-independent case.

In this work, we aim at delivering a fair splitting criterion termed \textit{SCAFF}: Splitting Criterion AUC for Fairness.
It allows for fair tree classifier learning using the threshold-independent performance and fairness measure of strong demographic parity.
Our result will be a \textit{fair tree classifier} learning algorithm which simultaneously (1) optimises for threshold-independent performance and strong demographic parity, (2) handles various multicategorical sensitive attributes simultaneously, (3) is tunable with respect to the performance-fairness trade-off during learning, and (4) extends to bagging and (gradient) boosting architectures.

The structure of the paper follows:
Sec.~\ref{prob_statement} expresses our problem statement formally;
Sec.~\ref{rel_work} discusses related work;
Sec.~\ref{method} elaborates our SCAFF method in detail;
Sec.~\ref{exps} describes our experiments;
Sec.~\ref{results} refers to our results; and
Sec.~\ref{conclusion} concludes and recommends research directions.

\section{Problem Statement}\label{prob_statement}
We consider the scenario in which a labelled dataset is intrinsically biased with respect to one or more sensitive attributes of which the values may be either binary or multicategorical.
Our task is to learn a fair predictive model from the biased data, such that future predictions are independent from the sensitive attribute(s).
We require that the definitions of model performance and fairness do not depend on a decision threshold set upon the output.
Since there is no unique solution in the trade-off between classification performance and fairness, the fair classification model must also be tunable in this regard.

Formally, consider a dataset $D$ with $n$ samples, $m$ features, and two classes.
Without loss of generality, assume the case in which a single binary sensitive attribute exists.
Let $X$, $Y$, and $S$ be the underlying variable distributions representing the feature space, classes, and sensitive attribute, respectively, from which the $n$ samples were drawn.
Accordingly, each sample may be represented as $(x_i, y_i, s_i)$, for $i = 1, 2, \dots, n$.

The goal of the fair learning algorithm is to learn the distribution for which the conditional $P(Y|X) \approx P(Y|X,S)$.
In practice, this amounts to learning from the data a mapping function $f: x \in X \to z \in Z$ where $Z$ represents the model output (i.e., classification score) upon which a threshold $t$ induces a class prediction, and under which the condition of strong demographic parity must be met,
$\forall t \in Z: P(Z \geq t|S_+) = P(Z \geq t | S_-)$,
while maximising for the threshold-independent classification performance $P[(Z|Y_+) \geq (Z|Y_-)]$.
The compromise between strong demographic parity and the corresponding maximal predictive performance must also be tunable.

\section{Related Work}\label{rel_work}
In this section, we discuss the concepts from the literature related to our work:
the measures of fairness (Sec.~\ref{fair_measures}), 
and the fair tree splitting criteria used towards fair tree classification learning (Sec.~\ref{tree_methods}).

\subsection{Measures of Fairness}\label{fair_measures}
Several fairness measures exist in the literature, which may be categorised as either (a) threshold-dependent or (b) threshold-independent.
The three most prevalent threshold-dependent measures are:
(1) demographic parity~\cite{dwork2012fairness};
(2) equal opportunity~\cite{corbett2018measure}; and
(3) equalised odds~\cite{hardt2016equality}.

First, \textit{demographic parity} is the condition under which each sensitive group (e.g. male/female) should be granted a positive outcome, at equal rates.
It is defined as the absolute difference between the proportion of positive class predictions $\hat{Y}_{+}$ in instances with a positive sensitive attribute value $S_{+}$ and instances with a negative sensitive attribute value $S_{-}$ and is formally given as
\linebreak
$|P(\hat{Y}_{+}|S_{+}) - P(\hat{Y}_{+}|S_{-})|$.
Second, the measure of \textit{equal opportunity} accounts for the predictive reliability within each sensitive group.
It is computed by taking the absolute difference of the true positive rate between the instance groups composed of the positive and negative sensitive attribute values
\linebreak
$|P(\hat{Y}_{+}|S_{+}, Y_{+}) - P(\hat{Y}_{+}|S_{-}, Y_{+})|$.
\noindent
Third, \textit{equalised odds} extends the previous definition by also incorporating the unreliability of predictions in the sensitive groups.
It is computed as the absolute difference between the equal opportunity and its corresponding false positive rate $|P(\hat{Y}_{+}|S_{+}, Y_{-}) - P(\hat{Y}_{+}|S_{-}, Y_{-})|$.

Albeit computationally different, the three measures share at least one common aspect:
the output of the classification model must be binary; i.e., a decision threshold must be placed upon the continuous output which induces the class prediction.
As a result, a problem arises when applying these measures towards learning a fair classifier.
By being threshold-dependent, these measures of fairness are limited to being exclusively reliable for the specific threshold which produces the class prediction: there is no guarantee that fairness holds for different threshold values.
In practice, when learning several fair classifiers for real-world applications, (i.e., hyperparameter optimisation), the selection of the final classification model should not be dependent on any arbitrary threshold, as fairness should be maintained throughout.
Rather, the decision threshold should only be placed a posteriori,
according to the performance requirements of the end user (e.g., precision vs. recall) whilst incurring the minimum impact over fairness.

The notion of threshold-dependent demographic parity has been extended to the threshold-independent case, termed the \textit{strong demographic parity} condition, introduced in~\cite{jiang2020wasserstein}.
It takes into account the continuous output of the model, such that the ordering of the output should be independent of the sensitive groups.
It is computed as the absolute difference between the following probabilities
\linebreak
$|P[(Z|S_+) \geq (Z|S_-)] - P[(Z|S_+) < (Z|S_-)]|$.
However, the aforementioned work only considered the implementation of strong demographic parity for the logistic regression case.
This impacts applicability since state-of-the-art non-linear models cannot be learned which directly optimise towards the strong demographic parity condition. 
We therefore focus on expanding the implementation of strong demographic parity towards non-linear models, specifically to tree-based architectures.

\subsection{Fair Tree Splitting Criteria}\label{tree_methods}
One clear advantage of tree learning algorithms is that they may be designed with any arbitrary splitting-selection criterion.
The criterion does not have to be differentiable, as long as it is computationally tractable.
A second advantage of tree frameworks over other architectures is their verified performance within different domains, making them a state-of-the-art solution to classification problems~\cite{zabihi2017detection, dogru2018traffic, angenent2020large}.

The practice of learning fairness-aware tree classifiers is directly linked to the splitting criterion used to construct the tree structure.
Within the fair tree literature, we recommend the works by Kamiran et al.~\cite{kamiran2010discrimination} and Zhang and Ntoutsi~\cite{ijcai2019-205}, in which different approaches are used to measure classification performance and fairness.
The measures are then jointly used as splitting criteria during training to select the best split.

In the work by Kamiran et al., the authors propose to address the fair splitting criterion problem, in which \textit{discrimination} is defined in terms of the threshold-dependent demographic parity.
They do so by extending the concept of information gain in traditional classification towards the sensitive attribute.
Given a set of data $D$, a split is evaluated in terms of the information gain with respect to the class label:
\begin{equation}
    \mathit{IG}_Y = H_{Y}(D) - \sum_{i=1}^{k} \frac{|D_i|}{|D|} \cdot H_{Y}(D_i) \text{,}
\end{equation}
and the information gain with respect to the sensitive attribute, given by:
\begin{equation}
    \mathit{IG}_S = H_{S}(D) - \sum_{i=1}^{k} \frac{|D_i|}{|D|} \cdot H_{S}(D_i) \text{,}
\end{equation}
where $H_Y$ and $H_S$ denote the entropy with respect to the class label and the sensitive attribute, respectively, and $D_i, i=1, \dots, k$ denotes the partitions of $D$ induced by the split under evaluation.
Both information gains are then merged to produce two distinct compound splitting criteria by either:
(1) subtracting $\mathit{IG}_Y$ by $\mathit{IG}_S$, hereinafter termed $\text{Kamiran}_{\text{Sub}}$,
or
(2) dividing $\mathit{IG}_Y$ by $\mathit{IG}_S$, hereinafter denoted as $\text{Kamiran}_{\text{Div}}$.
Although this work was fundamental in establishing fair tree-learning frameworks, it is limited in scope since fairness is only considered as the threshold-dependent demographic parity.

In their work, Zhang and Ntoutsi propose FAHT: a fairness-aware Hoeffding tree.
Although the method was developed with online streaming classification as its focus with constant tree-structure updates, the splitting criterion developed may be generally applicable.
Similar to the method of Kamiran et al., the FAHT approach relies on a compound criterion composed of a class label part and a sensitive attribute part and addresses demographic parity.
\linebreak
Both works use the same class label information gain $\mathit{IG}_Y$.
However, the fairness component is computed differently between them.
Zhang Ntoutsi define the fairness gain $\mathit{FG}$ of a split as a function of the measured discrimination $Disc(D)$ of a set of data, computed as:
\begin{equation}
    \mathit{FG} = Disc(D) - \sum_{i=1}^{k} \frac{|D_i|}{|D|} \cdot Disc(D_i) \text{.}
\end{equation}
Here, the discrimination is defined as the demographic parity of the system $|P({Y}_{+}|S_{+}) - P({Y}_{+}|S_{-})|$.
The FAHT splitting criterion is then defined as:
\begin{equation}\label{faht}
    \text{FAHT} =  
    \begin{cases}
        \mathit{IG}_Y,& \text{if } \mathit{FG}=0\\
        \mathit{IG}_Y \cdot \mathit{FG},              & \text{otherwise}
    \end{cases} \text{.}
\end{equation}

These proposed fair tree approaches present some limitations, three of which deserve to be named in particular:
(1) the construction processes were developed with only threshold-dependent fairness in mind;
(2) both implementations only address a single binary sensitive attribute; and
(3) there exists no performance-fairness trade-off tuning parameter built into the splitting criteria.
In the following section, we describe our proposed treed-based framework which lifts these limitations.


\section{Method}\label{method}
In this section we describe our proposed method.
It is a probabilistic tree learning framework which
(1) optimises for strong demographic parity,
(2) is tunable with respect to the performance-fairness trade-off, and
(3) addresses multiple multicategorical sensitive attributes simultaneously.
We begin by addressing how the measure of strong demographic parity is implemented in Sec.~\ref{fair_measure}.
In Sec.~\ref{split}, we provide our compound splitting criterion which incorporates a tunable parameter towards the trade-off between classification performance and fairness.
In Sec.~\ref{tree_construct}, we describe the tree construction process, reporting on how our method extends towards the multivariate and multicategorically valued sensitive attribute scenario.
A working Python implementation of our algorithm can be found in~\cite{repository}.

\subsection{Strong Demographic Parity}\label{fair_measure}
The strong demographic parity condition aims to minimise the difference in candidates from the sensitive groups among the selected candidates, regardless of any arbitrary decision threshold $t$.
The goal is to minimise the expression
$|P[(Z|S_+) \geq (Z|S_-)] - P[(Z|S_+) < (Z|S_-)]|$ from Sec.~\ref{fair_measures}.
The condition of strong demographic parity may be reached by learning the classifier function $f$ which randomly orders the samples towards the sensitive groups, while maximising for performance $P[(Z|Y_+) \geq (Z|Y_-)]$.

In machine learning, the ROC-AUC (hereinafter, AUC) is a measure which expresses the quality of a sample ordering with respect to a binary label, where a random order results in $\text{AUC}=0.5$.
We find the fair classifier $f$ by optimising for an AUC value of $0.5$ on the sensitive attribute.
In order to solve the optimisation problem, we aim at minimising the AUC with $S_+$ as the positive class, which we denote as $\text{AUC}_{S_+}$.
\linebreak
Since $\text{AUC}_{S_+}=0$ is also maximally unfair, we define \textit{sensitive AUC} ($\text{AUC}_\text{S}$) ---$f_S$ from Sec.~\ref{prob_statement}--- as follows: 
\begin{multline}
    \text{AUC}_{\text{S}}(Z, S) = \max(1 - \frac{\sum_{i=1}^{s_+} \sum_{j=1}^{s_-} \sigma(Z_i, Z_j)}{s_+ \cdot s_-},\\\frac{\sum_{i=1}^{s_+} \sum_{j=1}^{s_-} \sigma(Z_i, Z_j)}{s_+ \cdot s_-}) \text{,}
\end{multline}
where
\begin{equation}
    \sigma(Z_i, Z_j) = 
    \begin{cases}
        1,               & \text{if } Z_i > Z_j\\
        \frac{1}{2},     & \text{if } Z_i = Z_j\\
        0,               & \text{otherwise}
    \end{cases} \text{.}
\end{equation}
Here, $s_+$ and $s_-$ are the number of all instances $S_{+}$ and $S_{-}$ respectively, and $Z_i$ and $Z_j$ represent the $Z$ output scores associated with each corresponding instance.
The $\max$ operator bounds the range of values to $[0.5, 1]$.
A completely biased classifier has $\text{AUC}_\text{S}$ of $1$, and $0.5$ indicates complete fairness (i.e., strong demographic parity of $0$).

\begin{figure*}[t!]
    \centering
    \includegraphics[width=\textwidth]{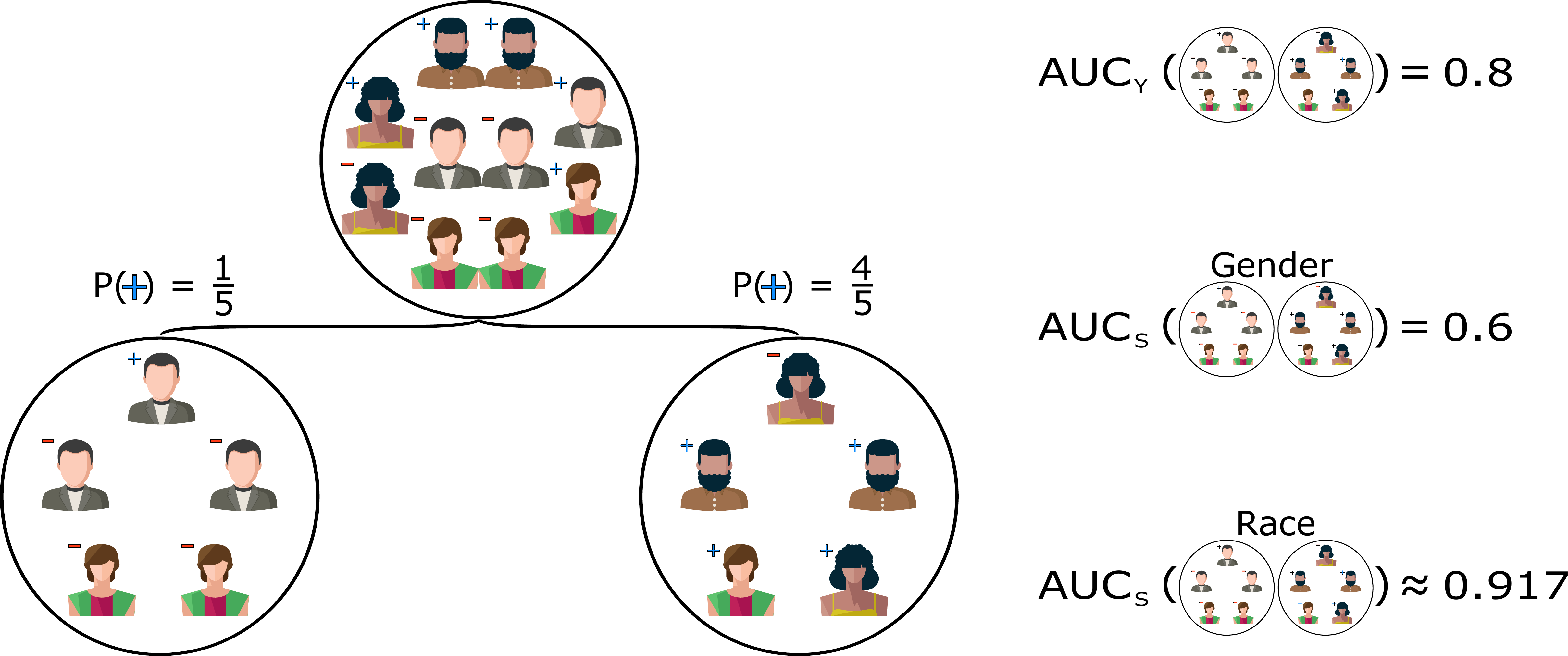}
    \caption{Computing necessary AUC values for split evaluation.
    Instances pertaining to a node are assigned its respective probability as their $Z$ scores.
    Considering orthogonality $\Theta=0.5$, then
    $SG=(0.5\cdot0.8 - 0.5\cdot0.6) - (0.5\cdot0.5 - 0.5\cdot0.5)=0.1$
    towards gender and, for race it follows
    $SG=(0.5\cdot0.8 - 0.5\cdot0.917) - (0.5\cdot0.5 - 0.5\cdot0.5)\approx-0.059$.}
    \label{fig:split}
\end{figure*}


\subsection{Splitting Criterion AUC for Fairness}\label{split}
The target performance measure should meet at least three requirements:
(1) it should be threshold-independent, as stated in our problem statement (Sec.~\ref{prob_statement});
(2) its values should be in the same range of the fairness measure, such that the tuning of the performance-fairness trade-off becomes intuitive for the end-user (i.e., practitioner), providing a simple interface to settle the fairness conditions of the use-case or application; and
(3) it should be computationally tractable, so that it can be applied to evaluate each candidate split.
One measure which satisfies all these requirements is
the standard classification AUC metric~\cite{fawcett2006introduction}, subsequently termed $\text{AUC}_\text{Y}$,
with complexity $O(n\cdot \log(n))$.

The objective becomes finding a split which maximises $\text{AUC}_\text{Y}$ (towards $\text{AUC}_\text{Y}=1$), while minimising $\text{AUC}_\text{S}$ (towards $\text{AUC}_\text{S}=0.5$).
Moreover, 
we propose an \textit{orthogonality} parameter $\Theta \in [0,1]$ which we incorporate into our splitting criterion.
The $\Theta$ parameter regulates the splitting criterion score towards either classification performance ($\Theta=0$) or fairness ($\Theta=1$).
Accordingly, for the simplest fair classification problem given instance scores $Z$, class label $Y$, and sensitive attribute $S$, we define SCAFF ---Splitting Criterion AUC for Fairness--- as:
\begin{multline}\label{eq_split}
     \text{SCAFF}(Z, Y, S, \Theta) =\\(1 - \Theta) \cdot \text{AUC}_\text{Y}(Z, Y) - \Theta \cdot \text{AUC}_\text{S}(Z, S) \text{.}
\end{multline}


\subsection{Tree Construction}\label{tree_construct}
As with any typical tree architecture, learning is done by selecting, at each step (i.e., depth), the split which optimises the splitting criterion score.
A split at some feature value partitions a node into two child nodes and is evaluated according to the $Z$ scores of the parent node and the new $Z'$ scores of the child nodes induced by that split.
The optimal split is the one which, across all possible feature value split points, maximises the splitting criterion score.

Given parent node scores $Z$ and child scores $Z'$ induced by a split, the $\text{SCAFF Gain}$ ($SG$) associated with that split is defined as:
\begin{equation}\label{eq:sg}
    SG = \text{SCAFF}(Z',Y,S, \Theta) - \text{SCAFF}(Z,Y,S, \Theta)\text{.}
\end{equation}
The split with maximal $SG$ across all evaluated splits is selected if and only if its corresponding $SG > 0$.
Otherwise, no splitting occurs and the parent node becomes a leaf node.
An example of SCAFF evaluation can be viewed in Fig.~\ref{fig:split}.
\linebreak
While we mention that $Z$ scores are defined as $P(Y_+)$ in a node, enabling bagging, other definitions are also viable.
For example, boosting techniques compute $Z$ by iteratively updating existing sample scores~\cite{hastie2009boosting}.
Our method extends to boosting since $SG$ relies on $Z$, regardless of its computation, whereas traditional fair tree learning algorithms do not, since no $Z$ scores are incorporated into the splitting criteria.
SCAFF extends to multivariate and multicategorical sensitive attributes, including intersectional factors (i.e., the combination of sensitive attributes)~\cite{intersectional} via a one-versus-rest (OvR) approach~\cite{tax2002using}.
The $\text{AUC}_\text{S}$ used in SCAFF is the maximum $\text{AUC}_\text{S}$ across all OvR, since no sensitive attribute should have priority over fairness.
Following from Fig.~\ref{fig:split}, the OvR $\text{AUC}_\text{S} = \max(0.6, 0.917) = 0.917$.


\section{Experiments}\label{exps}
For the description of our experiments, we begin by mentioning the datasets and how we used them (Sec.~\ref{datasets});
we then characterise the experimental setup deployed to (1) gather the performance and fairness values and (2) report on the relationship between the threshold-independent and threshold-dependent demographic parities (Sec.~\ref{exp_setup}).

We compared SCAFF against other fair splitting criteria by using benchmark fairness datasets.
Since the methods against which we compare our approach are neither suited for multivariate nor category-valued sensitive attributes, we focus on the single binary sensitive attribute case first.
We additionally experimented on a single dataset to explore how SCAFF handles multiple sensitive attributes simultaneously as well as multicategorical values.
Lastly, we tested the quantitative relationship of the strong demographic parity yielded by our method with the corresponding demographic parity at different decision-thresholds.
For reproducibility, our experiments are made available in~\cite{repository}.

\subsection{Datasets}\label{datasets}
Three binary classification datasets were used which have at least one sensitive attribute.
These are typical benchmark datasets used for fairness methods~\cite{quy2021survey}
Specifically, we employed the following:
(a) \textit{Bank} ($45,211$ instances, $50$ features) in which 
the sensitive attribute is the binary condition of age $\geq 65$
(b) \textit{Adult} ($45,222$ instances, $97$ features), where 
the sensitive attribute may be either (i) race $\in \{\text{white}, \text{non-white}\}$ or (ii) gender $\in \{\text{male}, \text{female}\}$; and
(c) \textit{Recidivism} ($6150$ instances, $8$ features) of which 
the sensitive attributes may be either (i) race $\in \{\text{white}, \text{non-white}\}$ or (ii) gender $\in \{\text{male}, \text{female}\}$.

For the binary sensitive attribute case, we considered each dataset-sensitive attribute configuration, making for a total of five different dataset configurations.
Two scenarios were further set in which the \textit{Adult} dataset was considered:
(i) the multiple sensitive attribute scenario such that both sensitive attributes (race and gender) were handled simultaneously; and 
(ii) the multicategorical sensitive attribute scenario in which
the intersectional attributes $\{$non-white female (NWF)$,$ non-white male (NWM)$,$ white female (WF)$,$ white male (WM)$\}$ were concurrently considered.

\subsection{Experimental Setup}\label{exp_setup}
To provide an adequate comparison between our splitting criterion and the state-of-the-art, we considered previous works in fair splitting criteria.
Specifically, we considered the works proposed by Kamiran et al.~\cite{kamiran2010discrimination} and Zhang and Ntoutsi~\cite{ijcai2019-205}.
For each dataset configuration, and for all method, the same $10$-fold cross validation was applied.

To measure classification performance and algorithm fairness, $\text{AUC}_{\text{Y}}$ (the accepted standard measure for classifier performance) and $\text{AUC}_{\text{S}}$ were used.
In line with our argumentation for using $\text{AUC}_{\text{S}}$ as a fairness measure in our splitting criterion, we apply it to measure the (un)fairness of the learned classifier.
The performance and fairness measures across test folds were averaged to produce a single value pair for each dataset, per method, and in our case for each value of orthogonality $\Theta$.
For all methods, the classification scores $Z$ of samples were computed as the $P(Y_+)$ of the terminal leaf node of a single tree.
To be able to achieve state-of-the-art performance, each method was deployed as a random forest (i.e., bagging)~\cite{breiman2001random}.
As such, the final classification score of a sample is the average $Z$ model output of all terminal nodes across the different trees generated.
Throughout all methods, the same set of hyperparameters was used, such as the number of trees ($500$), the maximum depth of each tree ($4$), and the random seed initialisation.
Bootstrapping, random feature selection, and continuous-feature discretisation were also applied, given their prevalence in real-world implementations of tree-based algorithms, such as XGBoost~\cite{xgboost}.
For our method, a range of $11$ values for $\Theta$ was used between $0$ and $1$.
For the implementation, see~\cite{repository}.

\begin{figure*}[b!]
    \centering
    \includegraphics[width=\textwidth]{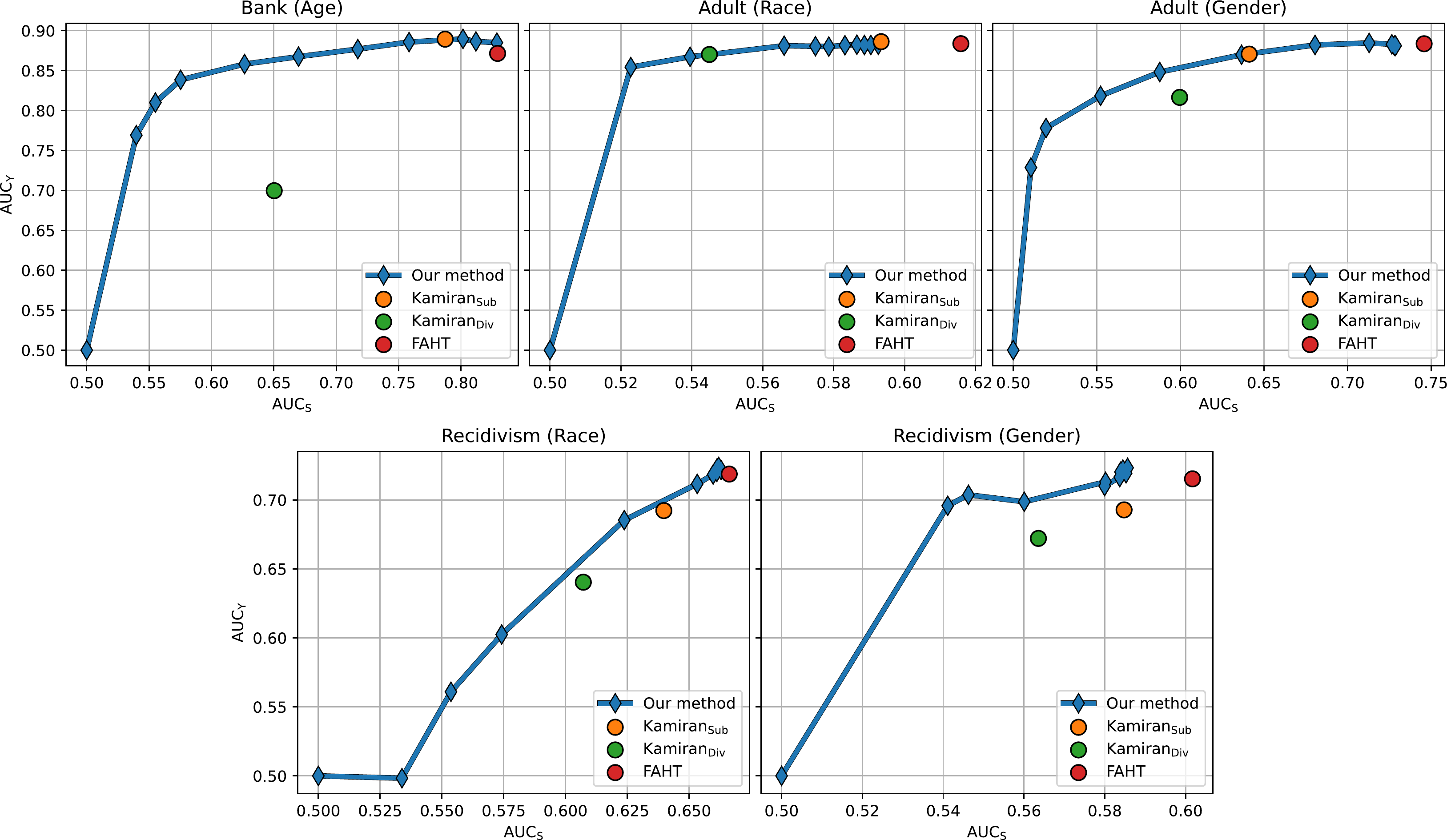}
    \caption{Model performance and fairness across methods, per dataset-sensitive attribute configuration.
    Measures of $\text{AUC}_{\text{S}}$ (horizontal axis) and $\text{AUC}_{\text{Y}}$ (vertical axis) are shown.
    For our method, each point from right to left indicates a value of $\Theta \in [0,1]$ in increasing order.}
    \label{fig:2d_results}
\end{figure*}

To relate the threshold-dependent and threshold-independent demographic parities, decision thresholds were applied to the classifier outputs of our method across different values of $\Theta$ for the different datasets.
The thresholds were considered as $9$ quantiles values between $0.1$ and $0.9$ of each test set output and, consequently, demographic parity ---defined in Sec.~\ref{fair_measures}--- was averaged over all folds.
\linebreak
Additionally, we measured at each decision threshold ---along $\Theta$ values--- the Pearson correlation coefficient~\cite{pearson}, and the respective null hypothesis p-values, between strong demographic parity (measured in $\text{AUC}_{\text{S}}$) and demographic parity.
The purpose is to check whether the behaviour of strong demographic parity across $\Theta$ in our method transfers to that of the demographic parity induced by any threshold.

\section{Results}\label{results}
In this section, we present the results of our experiments.
We begin by reporting on the classification performance and fairness obtained across our method and the competing approaches towards fair tree learning for the binary sensitive attribute configurations (Sec.~\ref{results_methods_binary}).
We follow with the performance and fairness for the non-binary case (Sec.~\ref{results_methods_multi}).
Finally, we show how strong demographic parity (measured in $\text{AUC}_{\text{S}}$) relates to demographic parity across different decision thresholds and values of orthogonality $\Theta$ (Sec.~\ref{relation_results}).

\subsection{Binary Sensitive Attribute}\label{results_methods_binary}
To regard the performance and fairness of all methods per dataset configuration, see Fig.~\ref{fig:2d_results}.
For our method, each point corresponds to a value of $\Theta \in [0,1]$.
Naturally, a $\Theta$ value of $0$ is equivalent to a traditional classifier (top-right).

In the horizontal axis, strong demographic parity is represented as $\text{AUC}_{\text{S}}$, while the vertical axis depicts the $\text{AUC}_{\text{Y}}$ classification performance.
Albeit differently-valued, the performance-fairness trade-off for each dataset-sensitive attribute pair (denoted at the top left of each graph) is consistent:
the greater the fairness (smaller values for $\text{AUC}_{\text{S}}$), resulting from increasingly greater values of $\Theta$, the lesser its classification performance (i.e., the fairness term acts as regularisation).
Unlike the other methods which output a single performance-fairness value (represented as a point), our SCAFF method produces a performance-fairness trade-off curve.
This is advantageous as it provides a way for practitioners to make an informed decision which suits their requirements.
The optimal fair classification solution should be to the top-left:
\textit{top} indicating high predictive performance, and
\textit{left} indicating low bias towards the sensitive attribute (in which a value of $0.5$ indicates a perfectly un-biased (or conversely, completely fair) classifier.
Noticeably, in \textit{Bank (Age)}, SCAFF was able to reduce $\text{AUC}_\text{S}$ by $0.2$ at a loss in performance of only $~0.02$.

Overall, our method consistently performs better in the combination of classification performance and fairness, allowing for a suitable target point.
It is a convincing result of
(1) the use of AUC in the splitting criterion and
(2) the flexibility of the $\Theta$ parameter.


\begin{figure*}[hb!]
    \centering
    \includegraphics[width=\textwidth]{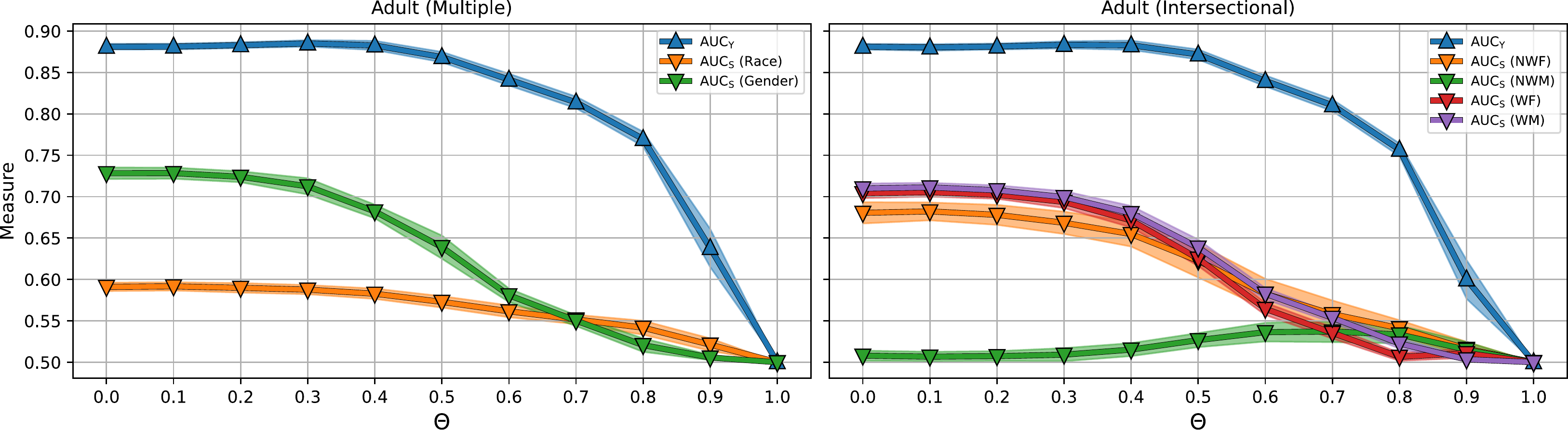}
    \caption{SCAFF classification performance and fairness for multiple and intersectional sensitive attributes of the \textit{Adult} dataset.
    Across different values of orthogonality $\Theta$ (horizontal axis), average and standard deviation of $\text{AUC}_{\text{Y}}$ and $\text{AUC}_{\text{S}}$ values (vertical axis) are shown.}
    \label{fig:2d_results_1}
\end{figure*}

\subsection{Multiple and Multicategorical Cases}\label{results_methods_multi}
\sloppy
We present in Fig.~\ref{fig:2d_results_1} the outcomes of the dataset configurations for multiple sensitive attributes ---\textit{Adult (Multiple)} in the left panel --- and multicategorical sensitive attribute values, considered as the intersectional values: \textit{Adult (Intersectional)} in the right panel.
For both panels, across different values of orthogonality $\Theta$ (horizontal axis), the classification performance $\text{AUC}_{\text{Y}}$ is shown in blue and the different $\text{AUC}_{\text{S}}$ are provided (vertical axis).
To the left, the $\text{AUC}_{\text{S}}$ for race and gender can be regarded;
to the right, the $\text{AUC}_{\text{S}}$ for each of the different intersectional sensitive attribute values are displayed: non-white female (NWF), non-white male (NWM), white female (WF), and white male (WM).

Noteworthily, SCAFF was able to reduce the bias towards both sensitive attributes simultaneously whilst maintaining adequate classification performance;
in particular at $\Theta = 0.7$, both race and gender $\text{AUC}_{\text{S}} = 0.55$ (a remarkably low bias value), and $\text{AUC}_{\text{Y}}$ is above $0.8$ indicating model prediction adequacy.
Similarly for \textit{Adult (Intersectional)} at the same orthogonality $\Theta=0.7$, our method was able to converge the bias of all sensitive attribute values to sensible values concurrently whilst maintaining proper classification performance.
These results show our proposed method is able to produce adequate classification models with regards to multiple and multicategorical sensitive attributes.

One limitation of our OvR approach to non-binary sensitive attributes is, however, regardable.
Since the OvR $\text{AUC}_{\text{S}}$ along multiple attributes or values is evaluated as its maximum (as described in Sec.~\ref{split}), there is no guarantee that all but the most biased attribute will have its fairness increased: regard the slight increase in bias for non-white males.
Yet, this characteristic of our approach also bounds the highest possible value of bias: along $\Theta$, the maximum value of $\text{AUC}_{\text{S}}$ is strictly monotonically decreasing.
The remark is further corroborated by the NWF, WF, and MF intersectional sensitive attributes, of which the curves behave in a nearly-identical manner along the different values of $\Theta$.
%

\subsection{Relationship with Demographic Parity}\label{relation_results}


Below, we describe the results of applying our method to the five dataset configurations for different values of $\Theta$, and measuring the corresponding (threshold-dependent) demographic parity at different decision thresholds.
The purpose is to determine if (1) threshold-independence extends across arbitrary decision thresholds, and (2) if the orthogonality parameter $\Theta$ induces a behaviour in demographic parity equivalent to the one in strong demographic parity.

\begin{figure*}[ht!]
    \centering
    \includegraphics[width=\textwidth]{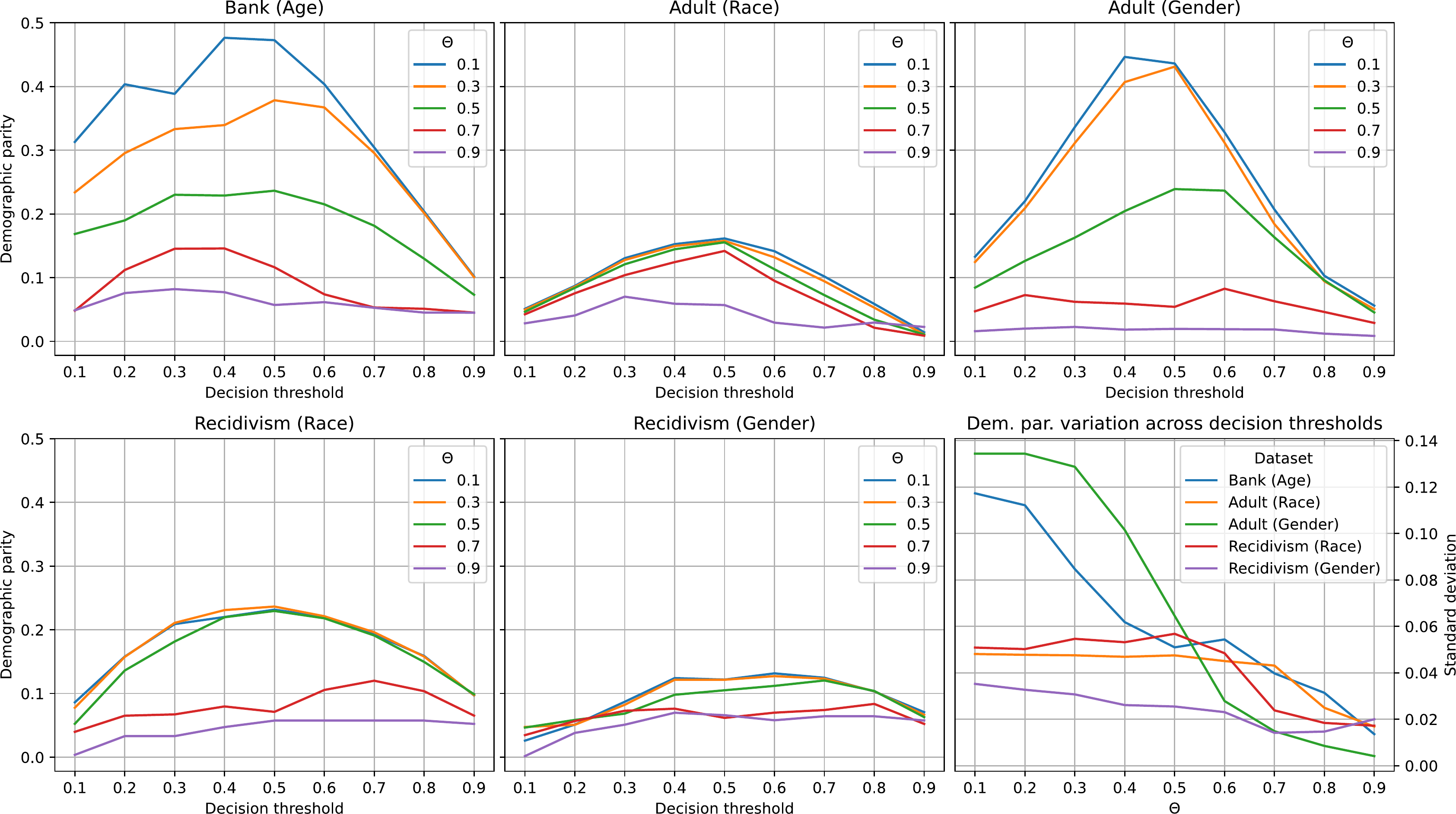}
    \caption{Effect of orthogonality over demographic parity.
    Decision thresholds are shown in the horizontal axis.~
    Values of mean demographic parity are represented in the vertical axis.
    Values of $\Theta$ are highlighted with different colours.
    Bottom-right panel: standard deviation (vertical axis) of demographic parity across decision thresholds, at different values of $\Theta$ (horizontal axis).
    }
    \label{fig:dem_parity}
\end{figure*}

In Fig~\ref{fig:dem_parity}, it is shown how for different decision thresholds (horizontal axis), the mean demographic parity (vertical axis) ---across all test folds--- behaves with different values of $\Theta$ (differently-coloured lines), for the five binary sensitive attribute dataset configurations.
An additional panel is provided (bottom-right), where for each value of $\Theta$ (horizontal axis), the variation of demographic parity across decision thresholds for each dataset is present.
Across all dataset configurations, and particularly noticeable in those with high demographic parity ---\textit{Bank (Age) and Adult (Gender)}--- the effect of $\Theta$ is generally the same.
As orthogonality increases, not only does demographic parity decrease, but so too does its spread (measured as standard deviation) across decision thresholds.
In other words, higher values of $\Theta$ translate to greater threshold-independence.
This is expected, as SCAFF directly optimises for threshold-independent measures.

\begin{table}[hb!]
    \centering
    \caption{Pearson correlation coefficients between strong demographic parity (measured as $\text{AUC}_{\text{S}}$) and demographic parity along $\Theta$, for different decision thresholds in the five dataset configurations.
    Bolded entries indicate a null hypothesis p-value $\leq 0.05$.}
    \begin{adjustbox}{width=\columnwidth}
    \begin{tabular}{rccccc}
    {} & \multicolumn{5}{c}{\textbf{Dataset}}\\
    \cmidrule{2-6}
    \multicolumn{1}{r|}{\textbf{Th}}    & Bank (A)          & Adult (R)         & Adult (G)         & Recid. (R)        & Recid. (G)        \\
    \midrule
    \multicolumn{1}{r|}{$0.1$}          & $\mathbf{0.983}$  & $\mathbf{0.963}$  & $\mathbf{0.994}$  & $\mathbf{0.937}$  & $\mathbf{0.839}$  \\
    \multicolumn{1}{r|}{$0.2$}          & $\mathbf{0.984}$  & $\mathbf{0.965}$  & $\mathbf{0.997}$  & $\mathbf{0.995}$  & $\mathbf{0.895}$  \\
    \multicolumn{1}{r|}{$0.3$}          & $\mathbf{0.993}$  & $\mathbf{0.971}$  & $\mathbf{0.994}$  & $\mathbf{0.987}$  & $\mathbf{0.968}$  \\
    \multicolumn{1}{r|}{$0.4$}          & $\mathbf{0.988}$  & $\mathbf{0.992}$  & $\mathbf{0.991}$  & $\mathbf{0.995}$  & $\mathbf{0.949}$  \\
    \multicolumn{1}{r|}{$0.5$}          & $\mathbf{0.997}$  & $\mathbf{0.988}$  & $\mathbf{0.995}$  & $\mathbf{0.990}$  & $\mathbf{0.973}$  \\
    \multicolumn{1}{r|}{$0.6$}          & $\mathbf{0.993}$  & $\mathbf{0.994}$  & $\mathbf{0.995}$  & $\mathbf{0.998}$  & $\mathbf{0.975}$  \\
    \multicolumn{1}{r|}{$0.7$}          & $\mathbf{0.984}$  & $\mathbf{0.979}$  & $\mathbf{0.984}$  & $\mathbf{0.991}$  & $\mathbf{0.992}$  \\
    \multicolumn{1}{r|}{$0.8$}          & $\mathbf{0.975}$  & $\mathbf{0.871}$  & $\mathbf{0.919}$  & $\mathbf{0.983}$  & $\mathbf{0.984}$  \\
    \multicolumn{1}{r|}{$0.9$}          & $\mathbf{0.941}$  & $0.267$           & $\mathbf{0.947}$  & $\mathbf{0.944}$  & $\mathbf{0.922}$  \\
    \end{tabular}\label{correlations}
    \end{adjustbox}
\end{table}

To grasp the relationship between strong and threshold-dependent demographic parities, regard Table~\ref{correlations}.
Each row depicts a decision threshold upon which demographic parity was computed, whereas a column indicates a dataset configuration.
Accordingly, a cell depicts the Pearson correlation coefficient between the two measures of fairness along the parameter $\Theta$, for a given decision threshold.
The coefficients represent how similar the behaviour between threshold-dependent and -~independent demographic parities is, induced by shifts in $\Theta$.
It is advantageous to maintain the behaviours similar, regardless of the selected threshold.

Noteworthily, bolded entries indicate a statistical significance of $\alpha=0.05$ towards the null hypothesis of no correlation.
Safe for a single outlying entry ---threshold $0.9$ in the \textit{Adult (Race)} configuration, in which the value of demographic parity is negligible--- all table entries are consistently high
and of statistical significance.
This shows that the effect of shifting the orthogonality parameter $\Theta$ is, in practice, identical for both types of demographic parity, validating our method with respect to threshold independence.


\section{Conclusion}\label{conclusion}
In the present work, we introduced SCAFF: the Splitting Criterion AUC for Fairness.
By doing so, we proposed a learning algorithm which simultaneously (1) optimises for threshold-independent performance ---ROC AUC--- and fairness ---strong demographic parity--- (2) is able to handle various multicategorical sensitive attributes simultaneously, (3) is tunable with respect to the performance-fairness trade-off during learning via an orthogonality parameter $\Theta$, and (4) is easily extendable to bagging and (gradient) boosting architectures.

We empirically validated our method through extensive experimentation.
Within our experiments with real datasets, we showed that our approach outperformed the competing state-of-the-art criteria methods, not only in terms of predictive performance and model fairness, but also by its capability of handling multiple sensitive attributes simultaneously, of which the values may be valued multicategorically.
Moreover, we demonstrated how the behaviour of strong demographic parity induced by our method extends to the threshold-dependent demographic parity.

As future work, we recommend
to extend the current framework from learning classification problems towards other learning paradigms.

\clearpage

{\small
\bibliographystyle{ieee_fullname}
\bibliography{main}
}

\end{document}